\documentclass[runningheads]{llncs}
\usepackage{graphicx}

\usepackage{tikz}
\usepackage{comment}
\usepackage{amsmath,amssymb} 
\usepackage{color}

\usepackage[accsupp]{axessibility}  

\usepackage{url}

\usepackage{multirow}
\usepackage{booktabs}
\usepackage{xcolor,colortbl}
\definecolor{Gray}{gray}{0.9}
\definecolor{citecolor}{RGB}{34,139,34}
\definecolor{grayDark}{gray}{0.95}
\definecolor{grayLight}{gray}{0.98}
\definecolor{darkgreen}{rgb}{0.10, 0.55, 0.10}
\newcommand\green[1]{\textcolor{darkgreen}{#1}}

\begin{document}
\pagestyle{headings}
\mainmatter
\def\ECCVSubNumber{5451}  

\title{Exploring Disentangled Content Information for Face Forgery Detection} 

\titlerunning{Exploring Disentangled Content Information for Face Forgery Detection}
\author{
    Jiahao Liang\inst{1} \and
    Huafeng Shi\inst{2} \and
    Weihong Deng\inst{1}\textsuperscript{*}
}
\authorrunning{J. Liang and H. Shi et al.}
%
\institute{Beijing University of Posts and Telecommunications \and
    SenseTime Research\\
    \email{\{jiahao.liang, whdeng\}@bupt.edu.cn, shihuafeng1@sensetime.com}}
\maketitle

\begin{abstract}
Convolutional neural network based face forgery detection methods have achieved
remarkable results during training, but struggled to maintain comparable
performance during testing.
We observe that the detector is prone to focus more on
content information than artifact traces,
suggesting that the detector is sensitive to the intrinsic bias of the
dataset, which leads to severe overfitting.
Motivated by this key observation,
we design an easily embeddable disentanglement framework for content
information removal,
and further propose a \textit{Content Consistency Constraint} ($\text{C}^2$C)
and a \textit{Global Representation Contrastive
Constraint} (GRCC) to enhance
the independence of disentangled features.
Furthermore, we cleverly construct two unbalanced datasets to investigate the impact of the content bias.
Extensive visualizations and experiments demonstrate that our framework can
not only ignore the interference of content information,
but also guide the detector to mine suspicious artifact traces
and achieve competitive performance.

\end{abstract}

\section{Introduction}
With the incredible success of deep learning, numerous techniques for
forgery have emerged, such as Deepfakes\cite{DeepFakes_github},
Face2Face~\cite{thies2016face2face},
and FaceSwap~\cite{FaceSwap_github}.
Due to the extremely low barriers and easy accessibility, forgery techniques
are gradually being misused~\cite{deepfake_porn,scam_a_ceo}.

To defend against, face forgery detection has attracted increasing
attention.
Early works~\cite{li2018ictu,yang2019exposing,matern2019exploiting,chugh2020not,haliassos2021lips} used hand-crafted facial
features (\emph{e.g.}, eyes blinking,
head poses, lip movements, \emph{etc.}) to capture some visual artifacts and
inconsistencies resulting from the forgery generation process.
Meanwhile, some works~\cite{ciftci2020fakecatcher,qi2020deeprhythm,liang2021identifying}
explored PPG signals representing heart rate
information.
Later, learning-based methods~\cite{yu2019attributing,ijcai2020-476,dang2020detection,chai2020makes,wang2020cnn}
have made significant progress.
Nevertheless, these methods are vulnerable to image compression
or noise interference.
Frank \emph{et al.}~\cite{frank2020leveraging} found that,
compared to the time domain, mining forgery information in the frequency domain
can still maintain satisfactory results even under severe
compression~\cite{frank2020leveraging,qian2020thinking,masi2020two,liu2021spatial,li2021frequency}.

\begin{figure}[t]

	\centering
		\includegraphics[width=1\linewidth]{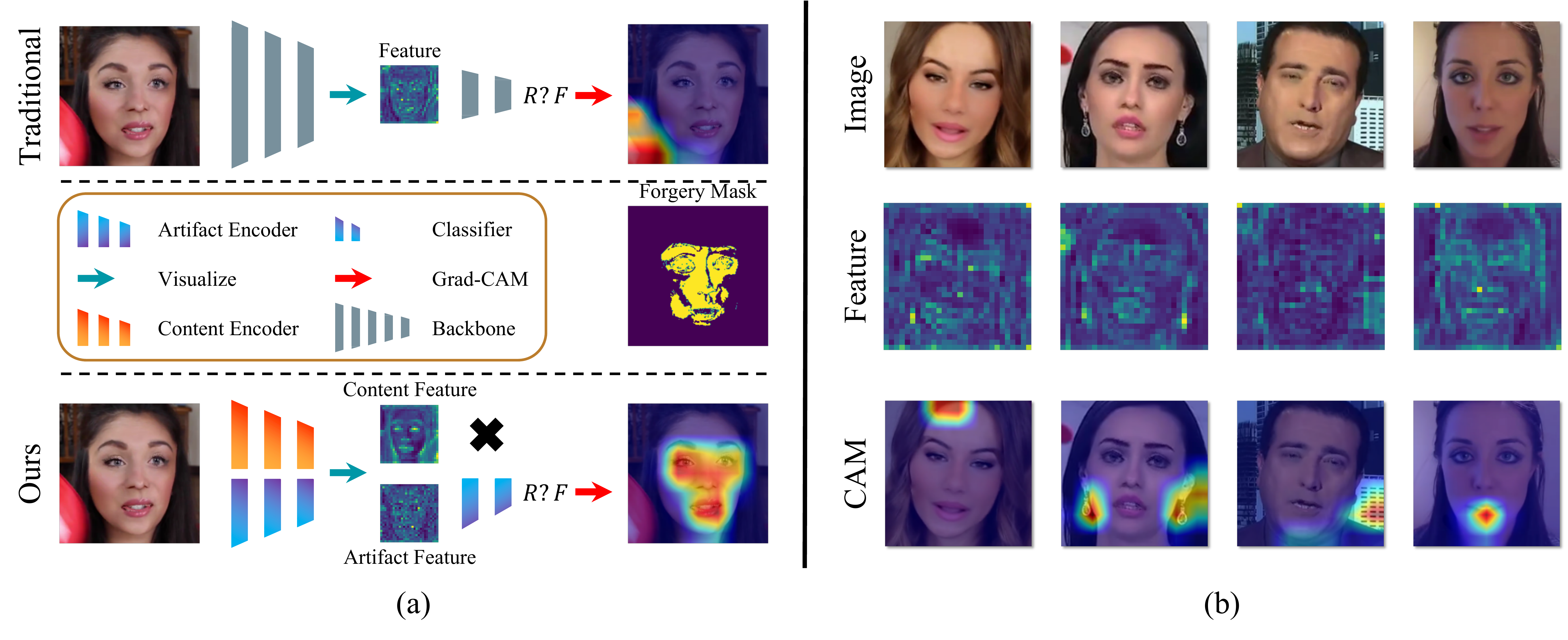}

	\caption{(a) Unlike the traditional methods (upper), we propose a disentanglement framework (lower)
			for content information removal. Grad-CAM~\cite{selvaraju2017grad} shows
			that the traditional detector is distracted by the red object,
			while our method still mine suspicious artifact traces and the activation region
			almost consistent with the mask.
			(b) Visualization of the image (first row), traditional detector's (Xception) features (second row)
			and Grad-CAM~\cite{selvaraju2017grad} (third row).}
	\label{figure:typical}
\end{figure}

We observe that most methods~\cite{zhao2021multi,qian2020thinking,masi2020two,qi2020deeprhythm}
perform admirably in in-dataset evaluations
but struggle to maintain comparable results in cross-domain evaluations, which
inspire us to conduct an in-depth analysis of the previous method.
Existing methods take it for granted that after proper training, the detector
will selectively grasp artifact traces as the basis for authenticity judgment.
However, the visualization (shown in Figure~\ref{figure:typical}~(b)) illustrates that
the feature of the detector remains recognizable content
clues, and the detector is prone to overfitting to
small local regions, or even focusing only on content information
outside the face region.

Based on this key observation, we conjecture that
detectors may no longer mine hard-to-capture artifact traces, and instead overfit
certain non-artifact (\emph{i.e.}, content) information,
thus leading to the failure of cross-domain evaluations.

Therefore, we propose an easily embeddable framework for disentangling content
features and artifact features, and only the disentangled
artifact
features
for face forgery detection, thus ignoring the interference of content
information.
A brief comparison between the traditional methods and our framework is sketched
in Figure~\ref{figure:typical}~(a).

However, most disentanglement methods~\cite{zhang2020face,liu2020disentangling,niu2020video}
consider only the completeness of
features, but do not explore the independence of disentangled features in-depth, which leads to
the failure of the face forgery detection (see Table~\ref{table:ablation}). To enhance it,
we propose a \textit{Content Consistency Constraint} ($\text{C}^2$C) to ensure that the disentangled features contain
the corresponding information and a \textit{Global
Representation Contrastive Constraint} (GRCC) to further ensure the purity of the disentangled features,
which helps our disentanglement framework to achieve competitive performance.
Furthermore,
we cleverly construct two unbalanced datasets
based on the FaceForensics++~\cite{rossler2019faceforensics++} to investigate the impact of content bias,
and further demonstrate that our framework can ignore the interference
of the content bias.
Notably, our framework is easily embeddable,
we embed some backbones into our framework for extensive evaluations and ablation experiments,
and experimental results demonstrate the effectiveness and generalization
capability of our framework in face forgery detection.

The contributions of this paper could be summarized as three-fold:
\begin{itemize}
\item To the best of our knowledge, we are the first to explore the
	impact of content information on the
	generalization performance of face forgery detection,
	and cleverly construct two unbalanced datasets to further investigate the impact of content bias,
	which brings a novel perspective for this field.
\item We design an easily embeddable disentanglement framework for content
	information removal,
	and further propose a \textit{Content Consistency Constraint} ($\text{C}^2$C)
	and \textit{Global Representation Contrastive Constraint} (GRCC) to enhance
	the independence of disentangled features.
\item Extensive visualizations and experiments demonstrate that our framework can
	not only ignore the interference of content information,
	but also guide the detector to mine suspicious artifact traces
	and achieve competitive performance in face forgery detection.
\end{itemize}

\section{Related Works}
\subsection{Forgery Detection}

Benefiting from the great progress of GAN,
forgery techniques, especially for faces, have been
incredibly advanced. To avoid its illegal use, researchers have explored forgery
detection extensively~\cite{li2020face,zhao2020learning,luo2021generalizing,sun2021domain,asnani2021reverse}.

Later, various learning-based methods~\cite{yu2019attributing,ijcai2020-476,dang2020detection,chai2020makes,wang2020cnn}
demonstrated significant improvements.
In addition, some works~\cite{afchar2018mesonet,liu2020global,chen2021local} suggested that shallow
local texture details and correlations between local regions of the face can
better reflect forgery information. However, almost all of
these CNN-based methods only utilize spatial domain information (\emph{i.e.},
RGB, YUV, HSV), and therefore the performance is sensitive to the quality and
distribution of the dataset. To counter it,
some works~\cite{frank2020leveraging,qian2020thinking,masi2020two,liu2021spatial,li2021frequency} transformed
images into the frequency domain by DCT transform and analyzed the frequency domain
statistics, achieving satisfactory results even with severe compression.
Recent attempts to boost the generalization of face forgery detection by
extending the activated attention region of the network.
Zhao \emph{et al.}~\cite{zhao2021multi} proposed multiple spatial attention
heads to guide the network focus on different local regions.
Wang \emph{et al.}~\cite{wang2021representative} encouraged detectors to dig
deeper into previously overlooked regions by masking the sensitive facial
regions.
Although these CNN-based methods significantly enhance the feature extraction
capability of the detector, due to the neglect of the content bias implied in
the features, the detector is trapped in the intrinsic bias
of the dataset, thus hindering the improvement of cross-domain generalization
performance.

\subsection{Disentangled Representation}
Disentangled representation learning is to decompose complex dimensional coupled
information into simple features with a strong distinguishing
ability~\cite{bengio2013representation}.
DR-GAN~\cite{tran2017disentangled} disentangled the face into identity and pose
features for synthesizing faces in arbitrary poses to aid in recognition.
Niu \emph{et al.}~\cite{niu2020video} proposed a cross-verified feature disentangling
strategy with robust multi-task physiological measurements.
Zhang \emph{et al.}~\cite{zhang2019gait} also adopted a similar
structure to disentangle pose and appearance features from gait videos.
In the field of face anti-spoofing, Zhang \emph{et al.}~\cite{zhang2020face} decompose
the facial image into content features and liveness features and introduced LBP map,
depth map as auxiliary supervision. Liu \emph{et al.}~\cite{liu2020disentangling}
proposed a new adversarial learning framework to separate the spoof trace into a
hierarchical combination of multi-scale patterns.
In this paper, we further propose a \textit{Content Consistency Constraint} ($\text{C}^2$C)
and \textit{Global Representation Contrastive Constraint} (GRCC) to enhance the independence of disentangled features.
And the disentanglement framework only serves as an underlying architecture,
a detailed ablation analysis can be found in Section~\ref{section:ablation}.

\section{Methods}

\subsection{Motivation}
Consider a forged image, which consists of artifact traces and content information,
where the content information can be subdivided into identity information and
background information.
The only difference between the forgery image and the real image is the presence of
artifact traces, which is the basis for the detector to determine the authenticity.

We observe that most detectors perform admirably
in in-dataset evaluations
but struggle to
maintain comparable results in cross-dataset evaluation.
For further exploration,
we visualize the features of the middle layer of the detector and the
Grad-CAM~\cite{selvaraju2017grad}. The visualization results
(see Figure~\ref{figure:typical}~(b)) illustrate that
the feature of the detector remains recognizable content
information, and the detector is prone to overfitting to
small local regions (DF, NT), or even focusing on content information
outside the face region (FS).

Based on this key observation, we conjecture that with the weak constraint of binary
labels alone,
detectors may no longer mine hard-to-capture artifact traces, and instead overfit
certain non-artifact information (\emph{i.e.}, content information),
thus failing in cross-dataset evaluations.

Therefore, we propose an embeddable disentanglement framework that disentangles
content and artifact features, and the artifact features are used for
forgery detection, thus eliminating the interference of content information.

\begin{figure*}[t]
	\centering
		\includegraphics[width=1\linewidth]{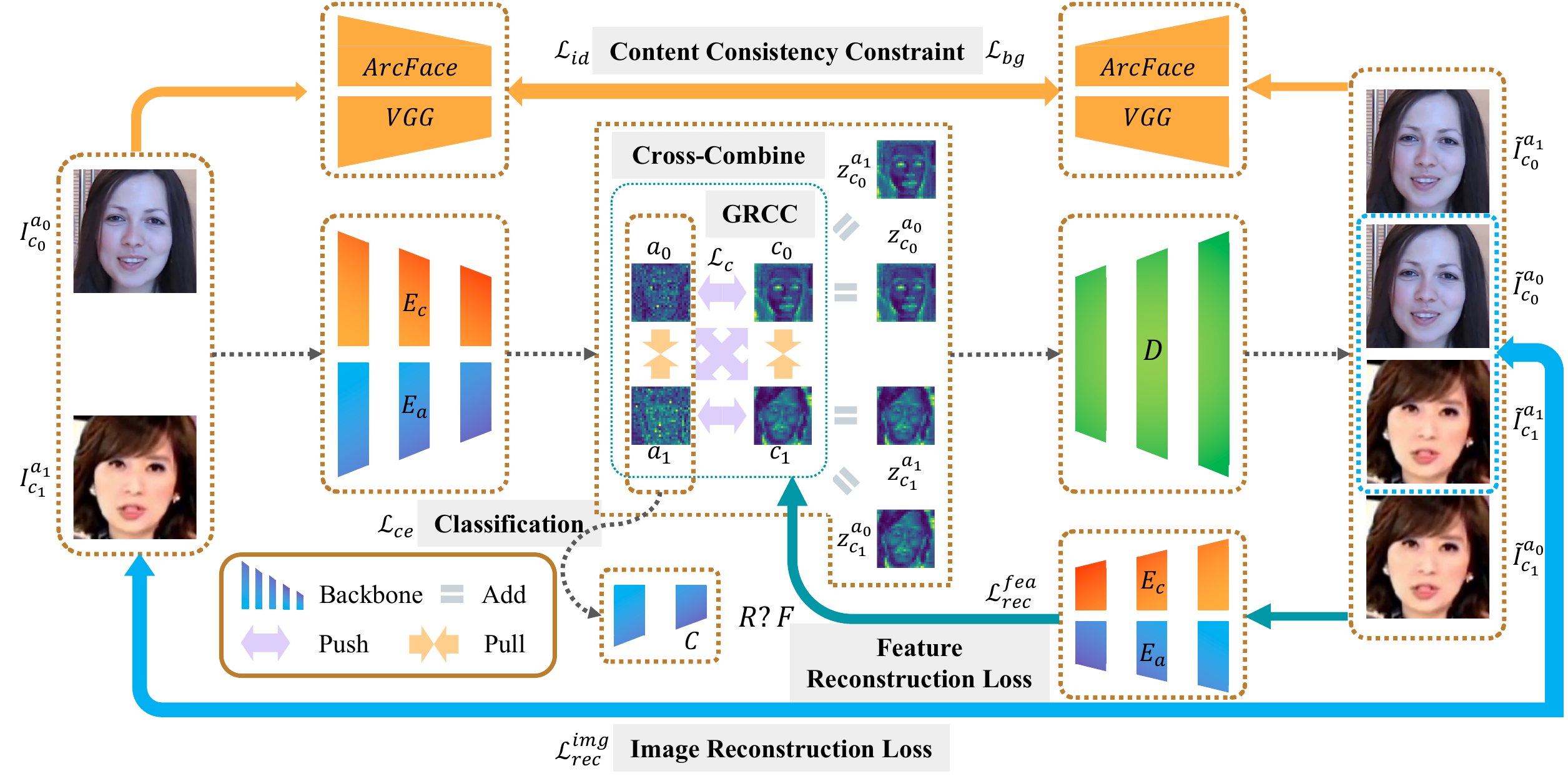}

	\caption{The overview framework of our method. The input of our network is
	a pair of images. First, we use artifact encoder $E_a$ and the content encoder
	$E_c$ to disentangle the content and artifact features, respectively. Then, we feed
	the artifact features $a_0$ and $a_1$ to the classifier $C$ for detection to compute
	$\mathcal{L}_{ce}$. Next, \textit{Global Representation Contrastive Constraint} (GRCC) is used
	to compute $\mathcal{L}_c$, and the artifact and content features are cross-combined to get latent representation $z_c^a$
	and then reconstruct the images. Finally,
the reconstruction loss $\mathcal{L}_{rec}^{img}$, $\mathcal{L}_{rec}^{fea}$ and the
\textit{Content Consistency Constraint} ($\text{C}^2$C) loss
 $\mathcal{L}_{id}$, $\mathcal{L}_{env}$ are calculated to ensure the completeness
and independence of the disentangled features.}
	\label{figure:framework}
\end{figure*}

\subsection{Basic Disentanglement Framework}
We assume that the high-dimensional latent representation of
an image consists of content and artifact features.
The main purpose is to disentangle them and use the
disentangled artifact
features for subsequent detection.

The disentanglement framework mainly consists of two independent encoders
$\mathit{E}_c$ and $\mathit{E}_a$, for extracting content and artifact features,
respectively, a decoder $\mathit{D}$ for the reconstruction of the images, and
a classifier $\mathit{C}$ for face forgery detection.
Among them, we use the front and back parts of the backbone as the artifact encoder $\mathit{E}_a$
and the classifier $\mathit{C}$, and the artifact encoder $\mathit{E}_a$ and the content
encoder $\mathit{E}_c$ have the same structure, but the parameters are not
shared.

Specifically, as shown in Figure~\ref{figure:framework},
with pairwise input images $I_{c_0}^{a_0}$,
$I_{c_1}^{a_1}$, where $a_0$, $a_1$ and $c_0$, $c_1$ denotes the
corresponding artifact and content features of the image, respectively.
It is worth noting that one of the images is real and the other one is fake.
We first use the content encoder $E_c$ and the artifact encoder
$E_a$ to get the content features
$c_0$, $c_1$ and the artifact features $a_0$, $a_1$, and the formula is as follow:
\begin{equation} \label{L_rec}
	c_i = E_c(I_{c_i}^{a_i}), a_i = E_a(I_{c_i}^{a_i}),
\end{equation}
where $i$ denotes the index of feature.

\noindent\textbf{Self-Reconstruction.}
Then element-wise addition is applied to the content and artifact features encoded
from the same image to obtain the high-dimensional latent representation
features of the image, \emph{i.e.}, $z_{c_i}^{a_i}=a_i+c_i$. Next,
$z_{c_i}^{a_i}$ is fed into the decoder $D$ to reconstruct
the corresponding original image $\widetilde{I}_{c_i}^{a_i}$, and the formula is as follow:
\begin{equation} \label{reconstruction}
	\widetilde{I}_{c_i}^{a_i} = D(z_{c_i}^{a_i}).
\end{equation}

\noindent\textbf{Cross-Reconstruction.}
Moreover, we \textbf{cross-combine} content and artifact features from different
images to obtain the high-dimensional latent representation features,
\emph{i.e.}, $z_{c_{1-i}}^{a_{i}}=a_i+c_{1-i}$.
Also, $z_{c_{1-i}}^{a_{i}}$
is fed into the decoder $D$ to reconstruct the image
$\widetilde{I}_{c_{1-i}}^{a_i}$,  and the formula is as follow:
\begin{equation} \label{cross_reconstruction}
	\widetilde{I}_{c_{1-i}}^{a_i} = D(z_{c_{1-i}}^{a_i}).
\end{equation}

\noindent\textbf{Reconstruction Loss.}
The decoder $D$ should effectively reconstruct the original image to ensure the
completeness of the high-dimensional latent representation feature,
so the image reconstruction loss is formulated as:

\begin{equation} \label{L_rec}
	\mathcal{L}_{rec}^{img} = \sum_{i=0}^1\left\|I_{c_i}^{a_i} - \widetilde{I}_{c_i}^{a_i}\right\|_1.
\end{equation}

Image reconstruction loss ensures that the reconstructed image and
the original image are consistent at the pixel level. In addition, the
encoded features of the reconstructed image should still be
consistent with the reconstructed features, so we introduce a feature
reconstruction loss:
\begin{equation}
	\small
	\begin{aligned}
		\mathcal{L}_{rec}^{fea}=&\sum_{i=0}^{1}(\left\|E_c(\widetilde{I}_{c_{i}}^{a_{i}})
		-c_i\right\|_1
		+ \left\|E_a(\widetilde{I}_{c_i}^{a_i}) - a_i) \right\|_1 \\
		+ & \left\| E_c(\widetilde{I}_{c_{1-i}}^{a_i}) - c_{1-i} \right\|_1
		+ \left\|E_a(\widetilde{I}_{c_{1-i}}^{a_i}) - a_i) \right\|_1).
	\end{aligned}
\end{equation}

\subsection{Enhanced Independence of Disentangled Features}
Although reconstruction loss can guarantee the completeness
of features for the combination of content and
artifact features.
However, there are still two elements that cannot be guaranteed:
(i) Whether the encoders can selectively disentangle features
(\emph{i.e.}, whether the disentangled features contain
the corresponding information).
(ii) Whether the disentangled features contain
\textbf{only} the corresponding information.
We are keenly aware that the key to successful disentangling lies in the
establishment of these two conditions, which is proved by subsequent
ablation study (Section~\ref{section:ablation}).
Unfortunately, none of the previous related
methods~\cite{zhang2020face,liu2020disentangling,niu2020video}
have explored the independence of features in depth.
We propose a \textit{Content Consistency Constraint} ($\text{C}^2$C) and a \textit{Global
Representation Contrastive Constraint} (GRCC) to further enhance the independence
of disentangled features.

\noindent\textbf{Content Consistency Constraint.}
In cross-reconstruction, content features should determine the background
and face ID information of the reconstructed image.
Specifically, the cross-reconstructed image $\widetilde{I}_{c_i}^{a_{1-i}}$
should have the same content attributes as the origin image $I_{c_i}^{a_i}$
that encodes the content features $c_i$.
As we mentioned before, content features consist of background and face ID,
so the \textit{Content Consistency Constraint} ($\text{C}^2$C) can be formulated as:
\begin{equation}
	\begin{aligned}
		\text{Content}(\widetilde{I}_{c_i}^{a_{1-i}}) &= \text{Content}(I_{c_i}^{a_i}), \\
		\Updownarrow \\
		\text{Identity}(\widetilde{I}_{c_i}^{a_{1-i}}) &= \text{Identity}(I_{c_i}^{a_i}), \\
		\text{Backgroud}(\widetilde{I}_{c_i}^{a_{1-i}}) &= \text{Backgroud}(I_{c_i}^{a_i}),
	\end{aligned}
\end{equation}
based on this prior condition, we
adopt the identity preservation loss $\mathcal{L}_{id}$ and the content perception loss $\mathcal{L}_{bg}$ to
preserve the content attributes of the cross-reconstructed images. It is
formulated as:
\begin{equation} \label{content_loss}
	\begin{aligned}
		\mathcal{L}_{id} & = 1 - \cos(\text{ArcFace}(\widetilde{I}_{c_i}^{a_{1-i}}), \text{ArcFace}(I_{c_i}^{a_i})), \\
		\mathcal{L}_{bg} & = \left\| \text{VGG}(\widetilde{I}_{c_i}^{a_{1-i}}) - \text{VGG}(I_{c_i}^{a_i}) \right\|_1,
	\end{aligned}
\end{equation}
where $\text{ArcFace}(\cdot)$ and $\text{VGG}(\cdot)$ represents a pretrained VGG network
and a pretrained
ArcFace network, respectively, $\cos(\cdot, \cdot)$
represents the cosine similarity of two vectors.
Here VGG(·) is considered to extract high-level semantic features,
and since artifacts are mainly concentrated in low-level texture details~\cite{liu2020global,zhao2021multi},
the extracted content features is pure and does not contain artifact information.

\noindent\textbf{Global Representation Contrastive Constraint.}
Artifact features and content features should be two fundamentally distinct spaces.
In other words, artifact features and content features can be regarded as two
different classes,
and the inter-classes feature distance should be much larger
than the intra-class feature distance.
Specifically, we regard the intra-class features as positive pairs and inter-class
features as negative pairs, and adopt the contrastive learning protocol to further
eliminate the possible overlap of content features and artifact features.
Inspired by \cite{liu2020global}, we take the Gram matrix of content and artifact
features as a global and distinctive representation:
\begin{equation} \label{gram}
\textbf{G}  =  (F_{i}^{T} F_{j})_{n\times n} =  
  \left[
   \begin{matrix}
  F^{T}_1 F_1 & \cdots & F^{T}_1 F_n \\
  \vdots & \ddots & \vdots \\
  F^{T}_n F_1 & \cdots  & F^{T}_n F_n \\
    \end{matrix}
  \right],
\end{equation}
where $F$ denotes the feature, and $n$ denotes the channel of the feature.
For feature distance measurement, we adopt the cosine
distance, where closer features render larger scores.
Finally, we take the advantage of the
InfoNCE~\cite{oord2018representation} to construct a
\textit{Global Representation Contrastive Constraint} (GRCC)
between the artifact and content features:
\begin{equation}
	\small
	\begin{aligned}
		\mathcal{L}_c = &-\log[\frac{\exp({d(\textbf{G}_{a_0}, \textbf{G}_{a_1})})}
	{\exp({d(\textbf{G}_{a_0}, \textbf{G}_{a_1})})
	+ \sum_{i=0}^1\exp({d(\textbf{G}_{a_i}, \textbf{G}_{c_{1-i}}}))}] \\
				  &- \log[\frac{\exp({d(\textbf{G}_{c_0}, \textbf{G}_{c_1})})}
	{\exp({d(\textbf{G}_{c_0}, \textbf{G}_{c_1})})
	+ \sum_{i=0}^1\exp({d(\textbf{G}_{a_i}, \textbf{G}_{c_{1-i}}}))}],
	\end{aligned}
\end{equation}
where $\textbf{G}_{a_i}$ and $\textbf{G}_{c_i}$ represent the flattened vector of the gram matrix of
$a_i$ and $c_i$, respectively, and $d(\cdot, \cdot)$ represents the cosine similarity.

\subsection{Overall Loss}
The final loss function of the training process is the weighted sum of the above
loss functions.
\begin{equation} \label{overall_loss}
	\mathcal{L} = \mathcal{L}_{ce}+\lambda_1\mathcal{L}_{rec}^{img}+\lambda_2\mathcal{L}_{rec}^{fea}
	+\lambda_3\mathcal{L}_{id}+\lambda_4\mathcal{L}_{bg}+\lambda_5\mathcal{L}_{c},
\end{equation}
where $\mathcal{L}_{ce}$ denotes the cross entropy loss, $\lambda_1$, $\lambda_2$,
$\lambda_3$, $\lambda_4$, $\lambda_5$ are the weights for balancing the loss.

\section{Experiments}
\subsection{Experimental Setting}
\noindent\textbf{Datasets.}
To validate the effectiveness of our method, we choose the most widely used benchmark
FaceForensics++ (FF++)~\cite{rossler2019faceforensics++} for training.
It contains 1 real sub-dataset and 4 fake sub-datasets,
\emph{i.e.}, Deepfakes (DF)~\cite{DeepFakes_github}, Face2Face (FF)~\cite{thies2016face2face},
FaceSwap (FS)~\cite{FaceSwap_github} and
NeuralTextures (NT)~\cite{thies2019deferred}. Each sub-dataset contains 1,000 videos, and
we follow the official standard by using 720 videos for training, 140 videos for
validation, and 140 videos for testing,
and we adopt the LQ version by default and specify the version otherwise.
Celeb-DF~\cite{li2020celeb} uses 59 celebrity interview videos on
YouTube as the original videos. In total, 590 real videos and 5,639
DeepFakes videos are included.

\noindent\textbf{Metrics.}
We apply the accuracy score (ACC), equal error rate (EER), and the area under the receiver operating
characteristic (ROC) Curve (AUC) as our evaluation metrics. For a comprehensive evaluation of
performance, we also report the true detection rate (TDR) for a given false detection
rate (FDR).

\noindent\textbf{Implementation Details.}
For data preprocessing, we only resize the facial images into a fixed size of $224\times224$.
For training, we set the size of the mini-batch to 128,
and the ratio of real and fake images to $1:1$.
We use Adam~\cite{kingma2014adam} as our optimizer with an initial learning rate
of 0.001 and a half
decay every 5000 iters. The maximum iters number is 30000. And we set $\lambda_1$
to $\lambda_5$ in Equation~\ref{overall_loss} as 1, 0.01, 1, 0.01 and 0.01.
All the code is based on the PyTorch framework and
trained with NVIDIA GTX 1080Ti.



\begin{table}[t]
	\centering
	\setlength\tabcolsep{3mm}
	\caption{In-Dataset evaluation (ACC~(\%)) on FF++ (LQ).
	We combine each forgery and real dataset in pairs to construct four
	sub-datasets, and evaluate the corresponding
	performance. AVG: the average performance of the four sub-datasets.
	Noting that results for some methods are from~\cite{qian2020thinking}.
	After embedding into our framework, all detectors achieve considerable
	performance gains and even outperform other methods.
}
	\setlength\tabcolsep{6mm}
	\scalebox{0.85}
	{
		\begin{tabular}{l|cccc|c}
			\specialrule{1pt}{1pt}{2pt}
			Method & DF & FF & FS & NT & AVG \\
			\hline
			Steg.Features~\cite{fridrich2012rich} & 67.00 & 48.00 & 49.00 & 56.00 & 55.00 \\
			LD-CNN~\cite{cozzolino2017recasting} & 75.00 & 56.00 & 51.00 & 62.00 & 61.00 \\
			C-Conv~\cite{bayar2016deep} & 87.00 & 82.00 & 74.00 & 74.00 & 79.25 \\
			CP-CNN~\cite{rahmouni2017distinguishing} & 80.00 & 62.00 & 59.00 & 59.00 & 65.00 \\
			MesoNet~\cite{afchar2018mesonet} & 90.00 & 83.00 & 83.00 & 75.00 & 82.75 \\
			$\text{F}^3\text{-Net}$~\cite{qian2020thinking} & 96.81 & 94.01 & \underline{95.85} & 79.36 & 91.51 \\

			\hline
    		\rowcolor{grayLight}
			Gram-Net~\cite{liu2020global} & 95.12 & 88.01 & 93.34 & 76.12 & 88.15 \\
    		\rowcolor{grayDark}
			+ Ours & \textbf{95.67} & \textbf{89.06} & \textbf{94.01} & \textbf{76.96} & \textbf{88.93} \\
			\hline
    		\rowcolor{grayLight}
			RFM~\cite{wang2021representative} & 95.42 & 91.24 & 93.60 & 79.83 & 90.02 \\
    		\rowcolor{grayDark}
			+ Ours & \textbf{95.92} & \textbf{92.27} & \textbf{93.97} & \textbf{80.14} & \textbf{90.58} \\
			\hline
    		\rowcolor{grayLight}
			ResNet-50~\cite{he2016deep} & 95.23 & 87.79 & 92.34 & 76.28 & 87.91 \\
    		\rowcolor{grayDark}
			+ Ours & \textbf{95.43} & \textbf{88.94} & \textbf{93.99} & \textbf{77.19} & \textbf{88.89} \\
			\hline
    		\rowcolor{grayLight}
			Xception~\cite{chollet2017xception} & 95.36 & 91.94 & 93.55 & 78.32 & 89.79 \\
    		\rowcolor{grayDark}
			+ Ours & \textbf{96.50} & \textbf{93.62} & \textbf{94.76} & \textbf{79.02} & \textbf{90.98} \\
			\hline
    		\rowcolor{grayLight}
			ResNest-50~\cite{zhang2020resnest} & 95.98 & 92.16 & 93.13 & 78.22 & 89.87 \\
    		\rowcolor{grayDark}
			+ Ours & \underline{\textbf{98.95}} & \underline{\textbf{94.32}} & \textbf{94.56} & \underline{\textbf{80.46}} & \underline{\textbf{92.10}} \\
			\specialrule{1pt}{1pt}{2pt}
		\end{tabular}%
	}
	\label{table:in_dataset}%
\end{table}%

\begin{table}[t]
	\centering
	\caption{
		Cross-Method evaluation (AUC~(\%)) on FF++ (C40). We adopt
		Xception~\cite{chollet2017xception}, which is widely used in face forgery detection,
		as a baseline for comparison on FF++. Specifically, we use one of the sub-datasets
		for training, and the rest for testing.
	}
	\setlength\tabcolsep{5mm}
	\scalebox{0.85}
	{
		\begin{tabular}{l|l|cccc|c}
    		\specialrule{1pt}{1pt}{2pt}
        \multirow{2}{*}{Train Set} & \multirow{2}{*}{Method} & \multicolumn{5}{c}{Test Set~(AUC(\%))} \\

			\cmidrule(lr){3-7}
													& & DF & FF & FS & NT & AVG \\
			\hline

			\multirow{2}*{DF} & \cellcolor{grayLight}Xception & \cellcolor{grayLight}99.21
							  & \cellcolor{grayLight}58.81 & \cellcolor{grayLight}64.79
							  & \cellcolor{grayLight}59.69 & \cellcolor{grayLight}70.63 \\
							  & \cellcolor{grayDark}+ Ours & \cellcolor{grayDark}\textbf{99.22}
							  & \cellcolor{grayDark}\textbf{60.18} & \cellcolor{grayDark}\textbf{68.19}
							  & \cellcolor{grayDark}\textbf{61.17} & \cellcolor{grayDark}\textbf{72.19} \\
			\hline

			\multirow{2}*{FF} & \cellcolor{grayLight}Xception & \cellcolor{grayLight}66.39
							  & \cellcolor{grayLight}95.40 & \cellcolor{grayLight}56.58
							  & \cellcolor{grayLight}57.59 & \cellcolor{grayLight}68.99 \\
							   & \cellcolor{grayDark}+ Ours & \cellcolor{grayDark}\textbf{67.13}
							   & \cellcolor{grayDark}\textbf{96.07} & \cellcolor{grayDark}\textbf{61.36}
							   & \cellcolor{grayDark}\textbf{59.98} & \cellcolor{grayDark}\textbf{71.14} \\
			\hline

			\multirow{2}*{FS} & \cellcolor{grayLight}Xception & \cellcolor{grayLight}80.00
							  & \cellcolor{grayLight}56.65 & \cellcolor{grayLight}94.55
							  & \cellcolor{grayLight}53.42 & \cellcolor{grayLight}71.16 \\
							   & \cellcolor{grayDark}+ Ours & \cellcolor{grayDark}\textbf{82.68}
							   & \cellcolor{grayDark}\textbf{56.77}& \cellcolor{grayDark}\textbf{94.76}
							   & \cellcolor{grayDark}\textbf{54.23} & \cellcolor{grayDark}\textbf{72.11}\\
			\hline

			\multirow{2}*{NT} & \cellcolor{grayLight}Xception & \cellcolor{grayLight}\textbf{69.94}
							  & \cellcolor{grayLight}\textbf{67.88} & \cellcolor{grayLight}57.59
							  & \cellcolor{grayLight}86.72 & \cellcolor{grayLight}\textbf{70.53} \\
							  & \cellcolor{grayDark}+ Ours & \cellcolor{grayDark}68.39
							  & \cellcolor{grayDark}65.40 & \cellcolor{grayDark}\textbf{58.34}
							  & \cellcolor{grayDark}\textbf{87.89} & \cellcolor{grayDark}70.01 \\
    		\specialrule{1pt}{1pt}{2pt}
		\end{tabular}
	}
	\label{table:cross_method}
\end{table}

\subsection{Evaluations} \label{Evaluations}
To evaluate our method comprehensively, in this section, we perform
in-dataset, cross-method and cross-dataset evaluation to demonstrate the
generalizability and robustness of our method.

\noindent\textbf{In-Dataset Evaluation.}
In-Dataset evaluation reflects the ability of the network to fit the distribution
of the dataset, as shown in Table~\ref{table:in_dataset}. In general, with the help
of our framework, the performance of both detectors and mainstream networks has
been improved in different degrees,
which fully proves the effectiveness and adaptability of our framework,
Among them, our methods (ResNest-50 + Ours) achieve the state of the art
on Deepfakes and NeuralTextures. Notably, for Deepfakes, we outperform
the $\text{F}^3\text{-Net}$~\cite{qian2020thinking} and baseline by 2.14\% and 2.97\% in terms of ACC score.
Although our best performance is still slightly worse than $\text{F}^3\text{-Net}$
on FaceSwap, it is understandable because our method does
not pursue a magical modification of the network architecture.

\noindent\textbf{Cross-Method Evaluation.}
Forgery techniques are constantly iterating, and we need to address not only existing
forgery methods, but also the most cutting-edge ones.
Table~\ref{table:cross_method} shows our method is superior to the baseline in most
cases, but the performance of both methods will drop greatly in cross-method
evaluation, which is inevitable, because the extremely strong feature
extraction capability of convolutional networks leads to the
overfitting of detectors.
Our method only mitigates the degree of overfitting to a certain extent.
but does not significantly improve the generalization performance.

\begin{table}[t]
	\centering
	\caption{Cross-Dataset evaluation on Celeb-DF (AUC~(\%) ) by training on FF++-DF (ACC~(\%)).
		Our method outperforms all the methods with the same backbone (Xception) and
		achieves the best performance with the backbone of ResNest-50.
}
	\setlength\tabcolsep{6mm}
	\scalebox{0.85}
	{
	\begin{tabular}{l|l|c|c}
    	\specialrule{1pt}{1pt}{2pt}
		BackBone & Method & FF++-DF~(Train) & Celeb-DF~(Test) \\
		\hline
		Xception & $\text{F}^3\text{-Net}$~\cite{qian2020thinking} & 97.97 & 65.17\\
		Efficient-B4 & Zhao \emph{et al.}~\cite{zhao2021multi} & - & 67.44 \\
		HRNet & Face X-ray~\cite{masi2020two} & - & 74.76 \\
		Xception & SPSL~\cite{liu2021spatial} & 96.91 & 76.88 \\
		- & Chen \emph{et al.}~\cite{chen2021local} & 98.84 & 78.26 \\
	    \hline
		\multirow{2}{*}{ResNet-18} & \cellcolor{grayLight}Gram-Net~\cite{liu2020global} & \cellcolor{grayLight}95.12
								   & \cellcolor{grayLight}67.14 \\
								   & \cellcolor{grayDark}+ Ours & \cellcolor{grayDark}\textbf{95.67}
								   & \cellcolor{grayDark}\textbf{74.94} \\
		\hline
		\multirow{2}{*}{Xception} & \cellcolor{grayLight}RFM~\cite{wang2021representative} & \cellcolor{grayLight}95.42
								  & \cellcolor{grayLight}67.21 \\
								  & \cellcolor{grayDark}+ Ours & \cellcolor{grayDark}\textbf{95.92}
								  & \cellcolor{grayDark}\textbf{74.44} \\
		\hline
		\multirow{2}{*}{ResNet-50} & \cellcolor{grayLight}ResNet-50~\cite{he2016deep} & \cellcolor{grayLight}95.23
								   & \cellcolor{grayLight}66.84 \\
								   & + \cellcolor{grayDark}Ours & \cellcolor{grayDark}\textbf{95.43}
								   & \cellcolor{grayDark}\textbf{74.71} \\
		\hline
		\multirow{2}{*}{Xception} & \cellcolor{grayLight}Xception~\cite{chollet2017xception} & \cellcolor{grayLight}95.36
								  & \cellcolor{grayLight}65.50 \\
								  & \cellcolor{grayDark}+ Ours & \cellcolor{grayDark}\textbf{96.50}
								  & \cellcolor{grayDark}\textbf{76.91}\\
		\hline
		\multirow{2}{*}{ResNest-50} & \cellcolor{grayLight}ResNest-50~\cite{zhang2020resnest} & \cellcolor{grayLight}95.98
									& \cellcolor{grayLight}68.00 \\
									& + \cellcolor{grayDark}Ours & \cellcolor{grayDark}\underline{\textbf{98.95}}
									& \cellcolor{grayDark}\underline{\textbf{82.38}} \\
    	\specialrule{1pt}{1pt}{2pt}
	\end{tabular}
	}
	\label{table:cross_dataset}
\end{table}

\begin{table}[t]
	\centering
	\caption{Ablation study on the FF++-DF and Celeb-DF. ``Basic'' represents
	the basic disentanglement framework.}
	\setlength{\tabcolsep}{6mm}
    \newcommand{\TableEntry}[2]{{#1}~\scriptsize{\green{(-#2)}}}
	\scalebox{0.85}
	{
		\begin{tabular}{l|ccc|cc}
			\specialrule{1pt}{1pt}{2pt}
			Method & Basic & $\text{C}^2$C & GRCC & FF++-DF & Celeb-DF \\
			\hline
			Xception & & & & 95.36 & 65.50 \\
			\rowcolor{grayLight}
			Variant A & \checkmark & & & 94.55 & 65.11 \\
			\rowcolor{grayDark}
			Variant B & \checkmark & \checkmark & & 95.73 & 70.08 \\
			\rowcolor{grayLight}
			Variant C & \checkmark & & \checkmark & 96.33 & 72.57 \\
			\rowcolor{grayDark}
			Variant D & \checkmark & \checkmark & \checkmark & \textbf{96.50} & \textbf{76.91} \\
			\specialrule{1pt}{1pt}{2pt}
		\end{tabular}
	}
	\label{table:ablation}
\end{table}

\begin{table}[t]
	\centering
	\caption{Results ($\Delta_{\text{AUC}}$~(\%)) of image- and feature-level data augmentation study.}
	\setlength{\tabcolsep}{5mm}
    \newcommand{\TableEntry}[2]{{#1}~\scriptsize{\green{(-#2)}}}
	\scalebox{0.85}
	{
		\begin{tabular}{l|ccc|ccc}
			\specialrule{1pt}{1pt}{2pt}
			\multirow{2}{*}{Method}& \multicolumn{3}{c|}{Augmentation} & \multicolumn{2}{c}{Dataset ($\Delta_{\text{AUC}}$~(\%))} \\
			\cmidrule(lr){2-4} \cmidrule(lr){5-6}
			& Erasing~\cite{zhong2020random} & H-Flip & Mixup~\cite{zhang2018mixup} & FF++-DF & Celeb-DF \\
			\hline
			\rowcolor{grayLight}
			Image & \checkmark &  &  & -2.13 & \textbf{+0.98} \\
			\rowcolor{grayDark}
			Feature & \checkmark &  &  & \textbf{+0.34} & +0.94 \\
			\hline
			\rowcolor{grayLight}
			Image &  & \checkmark &  & -0.10 & +0.23\\
			\rowcolor{grayDark}
			Feature &  & \checkmark &  & \textbf{+0.22} & \textbf{+3.62} \\
			\hline
			\rowcolor{grayLight}
			Image  &  &  & \checkmark & -0.85 & +3.01 \\
			\rowcolor{grayDark}
			Feature &  &  & \checkmark &  \textbf{-0.07} & \textbf{+3.57} \\
			\specialrule{1pt}{1pt}{2pt}
		\end{tabular}
	}
	\label{table:augmentation}
\end{table}

\noindent\textbf{Cross-Dataset Evaluation.}
Due to the differences in raw data and experimental details, there can be huge
gaps in the distribution between different datasets corresponding to even the same
method.
As shown in Table~\ref{table:cross_dataset}, regardless of the method,
the performance drops significantly
when testing on the Celeb-DF dataset, which implies that the difference in the
distribution of different datasets for the same method does exist.
With the assistance of our framework, the performance of each backbone on FF++
is slightly improved, but the improvement on Celeb-DF is significant.
Specifically, our method (ResNest-50+Ours) has a 14.38\% improvement on Celeb-DF,
while the improvement on FF++-DF is only 2.97\%.
Furthermore, our method (ResNest-50+Ours) outperforms the state-of-the-art
results (Chen \emph{et al.}~\cite{chen2021local}) by 4.12\% in terms of AUC score.
Among the methods using Xception as the backbone, our method
also surpasses others.

\subsection{Ablation Study} \label{section:ablation}
\begin{figure}[t]
	\centering
		\includegraphics[width=0.7\linewidth]{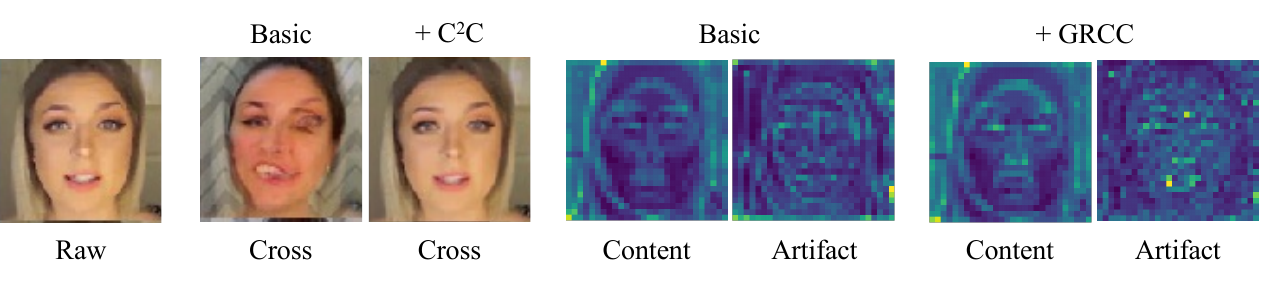}
	\caption{Visualization of the ablation study, which illustrates the
		impact of $\text{C}^2$C on the reconstructed images and GRCC on the
		disentangled features, respectively.``Raw'' represents the raw image,
		and ``Cross'' represents the cross-reconstruction image.}
	\label{figure:ablation}
\end{figure}

We perform several ablations to better understand the contributions of each component
in our method,
the experimental results and visualizations are shown in Table~\ref{table:ablation}
and Figure~\ref{figure:ablation}, respectively.

From the comparison of Variant A and Baseline, we can find that
the performance of face forgery detection does not increase but decreases
(0.81\%) by simply introducing the disentanglement framework.
Furthermore, we add \textit{Content Consistency Constraint} ($\text{C}^2$C) and
\textit{Global Representation Contrastive Constraint} (GRCC) separately,
with 4.97\% and 7.46\% improvement in terms of AUC, respectively, which
proves the effectiveness of the enhanced independence of disentangled features.
While the performance increases by 11.80\% after combining these two, which indicates
the two can play a mutually reinforcing role. Overall, $\text{C}^2$C and GRCC
play a dominant role as the key core of our method.

\begin{figure}[t]
	\centering
		\includegraphics[width=0.6\linewidth]{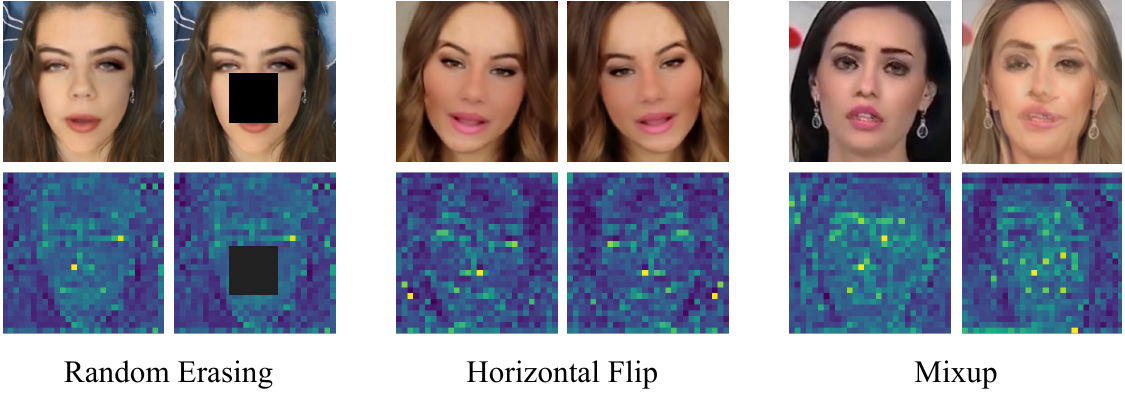}

	\caption{Visualization of the image- (1st row) and feature-level (2nd row) augmentation.}
	\label{figure:augmentation}
\end{figure}

\subsection{Augmentation Study of Disentangled Features}
Our framework first disentangles content features and artifact features from the images,
and then uses the artifact features for subsequent detection.
It is natural to guess that compared to image-level data
augmentation, directly performing data augmentation on artifact features may achieve
better performance. To validate it, we select common data augmentation methods such
as Random Erasing~\cite{zhong2020random}, Horizontal Flip, and Mixup~\cite{zhang2018mixup} to experiment,
the details of the augmentation are shown in Figure~\ref{figure:augmentation}.
It is worth noting that data augmentation is not used in other experiments.

It can be seen from Table~\ref{table:augmentation} that the performance improvement
in the cross-dataset evaluation is greater
than in-dataset evaluation. Our explanation is that in the in-dataset evaluation,
the distributions of the train and test sets are close, and data augmentation disrupts the
consistency of the distribution of the train and test set, resulting in little performance
improvement or even reduction. For cross-dataset evaluation, data augmentation can enhance
the diversity of the train set, and then pull the distribution between the train and test sets.
In addition, the performance improvement
of data augmentation at the feature-level is significantly better than that at the
image-level, which implies the effectiveness of our disentanglement framework.

\begin{figure}[t]
	\centering
		\includegraphics[width=1.0\linewidth]{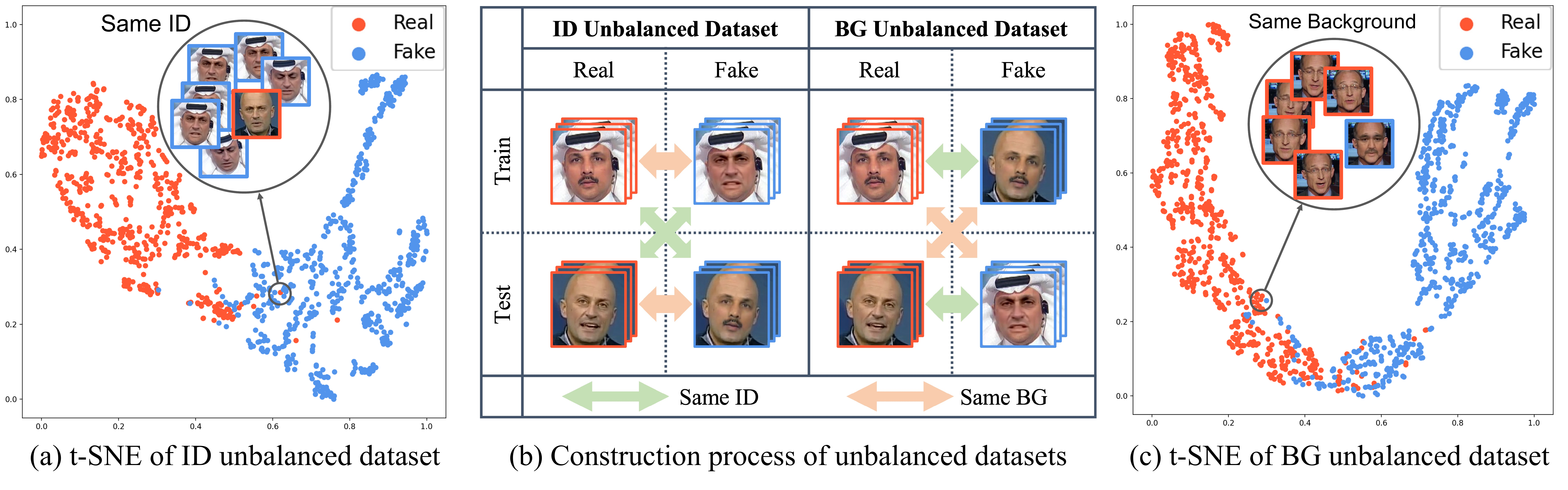}
	\caption{(b) The construction process of unbalanced datasets.
	(a)(c) t-SNE feature visualization of the
Xception network on the ID and BG unbalanced dataset.}
	\label{figure:tsne}
\end{figure}

\begin{table*}
	\centering
	\caption{Comparison of our framework with baseline methods on the identity and background unbalanced dataset.}
	\setlength\tabcolsep{2.5mm}
	\scalebox{0.85}
	{
		\begin{tabular}{l|cccc|cccc}
    		\specialrule{1pt}{1pt}{2pt}
        \multirow{2}{*}{Method}

			& \multicolumn{4}{c}{ID Unbalanced Dataset}
			& \multicolumn{4}{|c}{BG Unbalanced Dataset} \\
			\cmidrule(lr){2-5} \cmidrule(lr){6-9}
												 & ACC & AUC & EER & $\text{TDR}_{0.1}$
												 & ACC & AUC & EER & $\text{TDR}_{0.1}$ \\
			\hline
													   \cellcolor{grayLight}Gram-Net~\cite{liu2020global}
													   & \cellcolor{grayLight}89.85 & \cellcolor{grayLight}96.49
													   & \cellcolor{grayLight}10.14 & \cellcolor{grayLight}89.70
													   & \cellcolor{grayLight}79.84 & \cellcolor{grayLight}87.83
													   & \cellcolor{grayLight}20.19   & \cellcolor{grayLight}66.10 \\

													   \cellcolor{grayDark}+ Ours
													   & \cellcolor{grayDark}\textbf{94.80} & \cellcolor{grayDark}\textbf{99.48}
													   & \cellcolor{grayDark}\textbf{3.619} & \cellcolor{grayDark}\textbf{98.90}
													   & \cellcolor{grayDark}\textbf{94.00} & \cellcolor{grayDark}\textbf{98.93}
													   & \cellcolor{grayDark}\textbf{5.764} & \cellcolor{grayDark}\textbf{96.70} \\
			\hline
													  \cellcolor{grayLight}RFM~\cite{wang2021representative}
													  & \cellcolor{grayLight}90.34 & \cellcolor{grayLight}95.34
													  & \cellcolor{grayLight}9.232 & \cellcolor{grayLight}90.93
													  & \cellcolor{grayLight}85.09 & \cellcolor{grayLight}92.33
													  & \cellcolor{grayLight}14.89 & \cellcolor{grayLight}79.58 \\

													  \cellcolor{grayDark}+ Ours
													  & \cellcolor{grayDark}\textbf{95.49} & \cellcolor{grayDark}\textbf{99.11}
													  & \cellcolor{grayDark}\textbf{4.102} & \cellcolor{grayDark}\textbf{97.90}
													  & \cellcolor{grayDark}\textbf{95.02} & \cellcolor{grayDark}\textbf{98.34}
													  & \cellcolor{grayDark}\textbf{4.839} & \cellcolor{grayDark}\textbf{96.93} \\
			\hline
													   \cellcolor{grayLight}ResNet-50~\cite{he2016deep}
													   & \cellcolor{grayLight}89.61 & \cellcolor{grayLight}96.46
													   & \cellcolor{grayLight}10.35 & \cellcolor{grayLight}89.30
													   & \cellcolor{grayLight}80.88 & \cellcolor{grayLight}91.29
													   & \cellcolor{grayLight}17.17 & \cellcolor{grayLight}72.30 \\

													   \cellcolor{grayDark}+ Ours
													   & \cellcolor{grayDark}\textbf{95.39} & \cellcolor{grayDark}\textbf{99.54}
													   & \cellcolor{grayDark}\textbf{3.571} & \cellcolor{grayDark}\textbf{99.00}
													   & \cellcolor{grayDark}\textbf{94.46} & \cellcolor{grayDark}\textbf{98.76}
													   & \cellcolor{grayDark}\textbf{5.524} & \cellcolor{grayDark}\textbf{97.20} \\
			\hline
													  \cellcolor{grayLight}Xception~\cite{chollet2017xception}
													  & \cellcolor{grayLight}91.06 & \cellcolor{grayLight}96.91
													  & \cellcolor{grayLight}8.967 & \cellcolor{grayLight}91.50
													  & \cellcolor{grayLight}84.39 & \cellcolor{grayLight}92.42
													  & \cellcolor{grayLight}17.17 & \cellcolor{grayLight}78.10 \\

													  \cellcolor{grayDark}+ Ours
													  & \cellcolor{grayDark}\textbf{95.85} & \cellcolor{grayDark}\textbf{99.32}
													  & \cellcolor{grayDark}\textbf{3.434} & \cellcolor{grayDark}\textbf{99.10}
													  & \cellcolor{grayDark}\textbf{95.14} & \cellcolor{grayDark}\textbf{98.71}
													  & \cellcolor{grayDark}\textbf{4.762} & \cellcolor{grayDark}\textbf{97.00} \\
			\hline
														\cellcolor{grayLight}ResNest-50~\cite{zhang2020resnest}
														& \cellcolor{grayLight}89.89 & \cellcolor{grayLight}97.54
														& \cellcolor{grayLight}8.507 & \cellcolor{grayLight}92.70
														& \cellcolor{grayLight}81.28 & \cellcolor{grayLight}93.68
														& \cellcolor{grayLight}14.34 & \cellcolor{grayLight}79.60 \\

														\cellcolor{grayDark}+ Ours
														& \cellcolor{grayDark}\textbf{95.58} & \cellcolor{grayDark}\textbf{99.61}
														& \cellcolor{grayDark}\textbf{3.190} & \cellcolor{grayDark}\textbf{98.90}
														& \cellcolor{grayDark}\textbf{94.56} & \cellcolor{grayDark}\textbf{98.48}
														& \cellcolor{grayDark}\textbf{5.479} & \cellcolor{grayDark}\textbf{96.90} \\
    		\specialrule{1pt}{1pt}{2pt}
		\end{tabular}%
	}
	\label{table:extreme}%
\end{table*}%

\subsection{Investigation of Intrinsic Content Bias}
To investigate the impact of intrinsic content bias within the dataset on the
performance of face forgery detection,
we cleverly construct two unbalanced datasets based on the
FF++ dataset, the \textit{Identity Unbalanced} dataset and the \textit{Backgroud
Unbalanced} dataset.

We conduct comparative experiments on these datasets, and
the experimental results
are shown in Table~\ref{table:extreme}.
We can find that
the performance on the ID and BG unbalanced
datasets suffers a huge drop, which indicates that the existence of the intrinsic bias
does interfere with the optimization of the detector. In contrast,
our framework can maintain a high performance even on the unbalanced dataset by
stripping the content features and thus eliminating the interference of content bias.
Furthermore, compared with the ID unbalanced dataset, the performance degradation on the
BG unbalanced dataset is more serious.

For a more intuitive understanding of the impact of content bias,
we also visualize the t-SNE~\cite{van2008visualizing} feature spaces of the
Xception network on the ID unbalanced dataset (Figure~\ref{figure:tsne}~(a))
and BG unbalanced dataset (Figure~\ref{figure:tsne}~(c)).
We can observe that some samples with similar content information tend to cluster
together, in other words, the distance between some samples with similar content
information is much smaller than the distance between samples with similar forgery
methods, which reveals that the content bias
induce the detector to use content
information for discrimination instead of artifact traces.

\begin{figure}[t]
	\centering
		\includegraphics[width=1\linewidth]{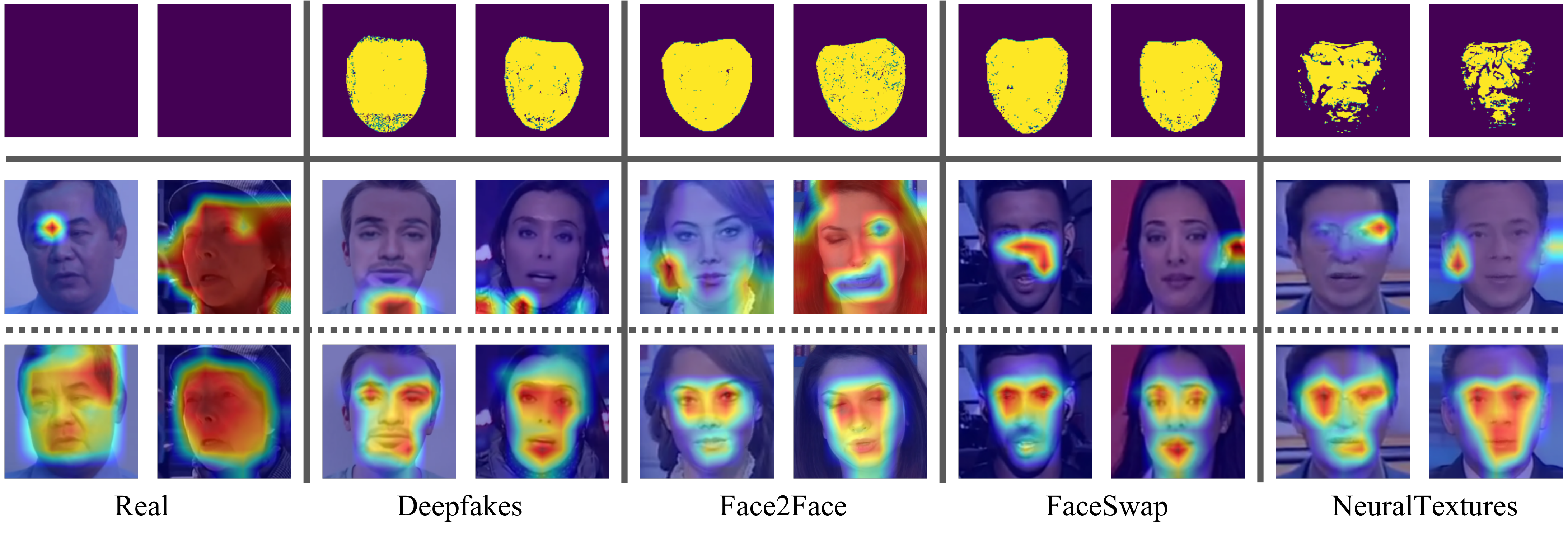}

	\caption{
		Visualization of forgery mask (first row), Xception's (second row)
		and ours (third row) Grad-CAM on five sub-datasets of FF++.
		The activation region of our method is comprehensive and
		almost consistent with the forgery mask.
	}
	\label{figure:visualize_cam}
\end{figure}

\begin{figure}[t]
	\centering
		\includegraphics[width=1\linewidth]{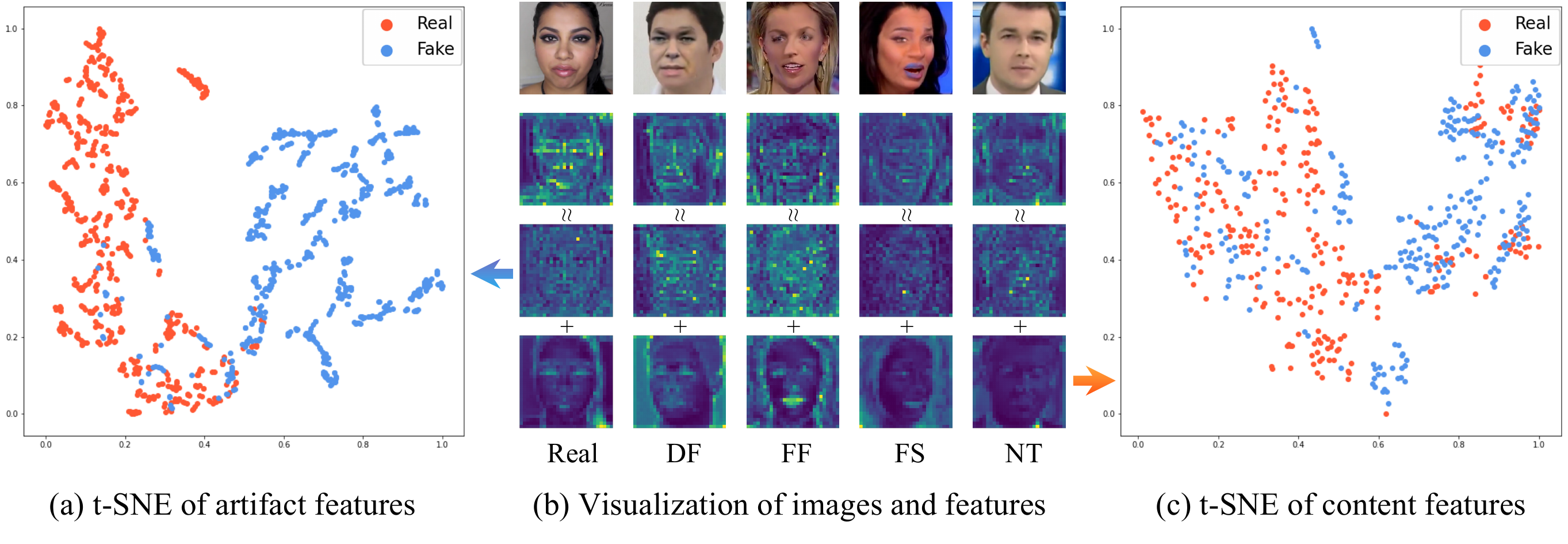}
	\caption{(b) Visualization of the image (first row), traditional detector's (Xception) features (second row),
		ours disentangled artifact (third row) and content features (fourth row).
		(a)(c) t-SNE visualization of artifact features and content features.
	}
	\label{figure:visualize}
\end{figure}

\subsection{Visualization}
To more intuitively demonstrate the effectiveness of our method, we visualize the
Grad-CAM~\cite{selvaraju2017grad}
of the baseline and our method, respectively, and the forgery mask, as
shown in Figure~\ref{figure:visualize_cam}. Grad-CAM shows that the baseline
is prone to overfitting to small local regions or focusing on content
noise outside the forgery region. In contrast, the activation region of
our method is comprehensive and
almost consistent with the forgery mask.
Such visualization results also explain the motivation of this paper:
without additional constraints, the detector has difficulty in mining
suspicious artifact regions thorough weak supervision of labels only,
and easily falls into content bias, which leads to overfitting or even
misleading the direction of optimization. Instead, our goal is to remove
the interference of content bias by an pre-disentanglement framework, and
guide the detector to mine suspicious artifact trace.

As shown in the Figure~\ref{figure:visualize}~(b),
traditional methods seek to allocate more attention to the face region, which improve
the fitting ability but also exacerbated the overfitting of content bias within
the dataset. Instead, we separate content features to eliminate misleading
content information, guide the detector to pay attention to suspicious
artifact traces, and strengthen the generalization
capability fundamentally. Furthermore, Figure~\ref{figure:visualize} (a)(c)
demonstrate that the disentangled artifact features are discriminative for
forgery detection, while the content features do not,
which also validates the validity of our motives.

\section{Conclusion}
In this paper,
we observe that
detectors may no longer mine hard-to-capture artifact traces, and instead overfit
certain content information,
thus leading to the failure of generalization,
which brings a novel perspective for face forgery detection.
Motivated by this key observation,
we design an easily embeddable disentanglement framework for content
information removal,
and further propose a \textit{Content Consistency Constraint}
($\text{C}^2$C) and a \textit{Global Representation Contrastive
Constraint} (GRCC) to enhance
the independence of disentangled features.
Furthermore, we cleverly construct two unbalanced datasets to
investigate the impact of the content bias.
Extensive visualizations and experiments demonstrate our framework
can not only ignore the interference of content bias
but also guide the detector to mine suspicious artifact traces
and achieve competitive performance in face forgery detection.

\noindent\textbf{Acknowledgments.}
This work was supported by National Key R\&D Program of China (2019YFB1406504).

\clearpage
\bibliographystyle{splncs04}
\bibliography{egbib}
\end{document}